\documentclass[smallextended,twocolumn]{svjour3} 
\smartqed
\usepackage[utf8]{inputenc}
\usepackage{subfigure}
\usepackage{graphicx}
\usepackage[numbers,sort]{natbib}
\usepackage{url}

\begin{document}
\title{The Many AI Challenges of Hearthstone}
\author{Amy K. Hoover \and Julian Togelius \and Scott Lee \and Fernando de Mesentier Silva}


\institute{Amy K. Hoover \at
              New Jersey Institute of Technology\\
              323 Dr Martin Luther King Jr Blvd\\
              Newark, NJ 07102 \\
              \email{ahoover@njit.edu}           
           \and
           Julian Togelius \at
              New York University\\
              6 MetroTech Center\\
              Brooklyn, NY 11201\\
              \email{julian@togelius.com}   
              \and
           Scott Lee \at
              Independent\\
              Irvine, CA 90066\\
              \email{randomperson2727@gmail.com}
              \and
          Fernando de Mesentier Silva \at
              Independent\\
              Rio de Janeiro, Brazil\\
              \email{fms2005@gmail.com}  
}


\maketitle

\begin{abstract}
Games have benchmarked AI methods since the inception of the field, with classic board games such as Chess and Go recently leaving room for video games with related yet different sets of challenges. The set of AI problems associated with video games has in recent decades expanded from simply playing games to win, to playing games in particular styles, generating game content, modeling players etc. Different games pose very different challenges for AI systems, and several different AI challenges can typically be posed by the same game. In this article we analyze the popular collectible card game Hearthstone (Blizzard 2014) and describe a varied set of interesting AI challenges posed by this game. Collectible card games are relatively understudied in the AI community, despite their popularity and the interesting challenges they pose. Analyzing a single game in-depth in the manner we do here allows us to see the entire field of AI and Games through the lens of a single game, discovering a few new variations on existing research topics.

\keywords{Artificial Intelligence \and Games \and Hearthstone \and Deckbuilding \and Gameplaying \and Player Modeling}

\end{abstract}

\section{Introduction}


For decades classic board games such as Chess, Checkers, and Go have dominated the landscape of AI and games research. Often called the ``drosophila of AI'' in reference to the \emph{drosophila} fly's significance in biological research, Chess in particular has been the subject of hundreds of academic papers and decades of research \citep{ensmenger:sss12}. At the core of many of these approaches is designing algorithms to beat top human players. However, despite IBM's Deep Blue defeating Garry Kasparov in the 1997 World Chess Championships and DeepMind's AlphaGo defeating Lee Sedol in the 2016 Google DeepMind Challenge Match \citep{silver:nature16}, such programs have yet to exhibit the general intelligence hoped for when such benchmarks were originally proposed. 


Since these victories, the AI and games community has gradually shifted focus to challenges in digital games. AI agents can outperform human players in simulated arcade games, sometimes beating them when a win condition is possible \citep{mnih:nature15,bellemare:jair13}. Agents and can also outperform humans in real time strategy games (RTSs) like Starcraft II \citep{deepmind:blog19} and multiplayer online battle arena games (MOBAs) like Defense of the Ancients 2 \citep{openai:dota18}. While there is value in designing algorithms to win (e.g. the popularity of minimax and alpha-beta pruning algorithms \citep{shannon:pmjs50,turing:ftt53} and Monte Carlo Tree Search (MCTS)  \citep{coulom:iccg06,silver:nature16}), like the drosphila fly necessarily shapes biological research it is possible that such focus limits the types of problems that can be solved in AI. Generally such agents only play the particular game they were built or trained to play.


However, both digital and analog games pose a variety challenges for which a variety of AI-based methods have been developed in response \citep{yannakakis:book18,lucas:tcaig09,browne:book11,cai:book07,kasparov:book17}
Research field of AI and games focuses not only on playing to win, but on modeling player behavior, modeling player experience, generating content and many other challenges. There are many reasons people play games beyond winning, and consequently many AI challenges present in any suitably rich game environment \citep{yee:cb06}. 


Rather than the typical approach of isolating particular challenges in artificial intelligence and solving them through exploration of a digital game, this paper instead discusses the multiple AI challenges posed by the popular collectible card game called Hearthstone. While there are other digitial collectible cards games like Gwent \citep{}, The Elder Scrolls:Legends \citep{}, and Clash of Clans \citep{}, Hearthstone is selected for its popularity with over 100 million players. While it is perhaps most common to consider such a game from a particular angle (e.g. playing to win it or modeling its players), this paper presents a kaleidoscopic view of the AI challenges the game presents and some of the current approaches to addressing them. We do not intend this to be an exhaustive survey of AI Challenges in Hearthstone, but to be representative; the AI challenges covered are likely to be closely related to some challenge already described for other games.

Hearthstone is a game rich in AI challenges and relatively unexplored, perhaps more than most games, given the many facets of the game. We also think that most of the research challenges identified here carry over to other collectible card games, and to some extent games that include elements of these games, such as deck building. Still, we believe that many other games have a rich diversity of AI challenges, far more than are usually considered, if you just look. This paper can therefore also serve as a paradigm for papers elucidating and cataloguing AI challenges in other types of games.

\section{The Challenges of Hearthstone}

Like many traditional board games, Hearthstone \citep{hearthstone:web18} is a two-player, turn-taking, adversarial game. However, unlike these games, it contains a large amount of stochasticity in play and partial information. The goal of this digital collectible card game is through playing different types of cards, decrease the opponents health from thirty to zero. Players initially choose one of nine different \emph{heroes}, which will determine the types of cards that the player can access. They then build decks of thirty cards from over 1900 which will be available in any given game. Each card costs the player a certain amount of a resource called mana, and has other attributes like attack, health, or spells it can cast.

While often combined by players in the way they discuss the game, at its core Hearthstone can be divided into two related challenges: 1) playing the game and 2) selecting decks to be played in matches. When describing their overall approach to others, players often refer to their deck archetype and hero, which is inherently packaged with heuristics to effectively play against other players. For example, the Odd Paladin is at the time of writing a powerful type of Paladin deck built around a particular card called Baku the Mooneater, which gets more powerful if all of the other cards in the deck cost an odd amount of mana to play. The deck favors an an aggressive gameplay strategy where the player focuses on destroying the enemy hero as quickly as possible rather than controlling the board or relying on clever card combinations. Each new expansion of Hearthstone changes the set of possible cards from which a player can build decks, that can in turn result in new sometimes more nuanced ways of play. So even when exploring playing to win, as the game is updated, so are the decks and strategies.  

The following sections enumerate and outline AI challenges in Hearthstone, including playing in Section \ref{sec:playing}, building decks in Section \ref{sec:decks}, helping human players learn to play and build decks in Section \ref{sec:assisting}, and helping designers build the game in Section \ref{sec:assistingdesigners}. These ideas are then combined to explore how the field of AI can be taught by approaches addressing these challenges in Hearthstone. But first we will outline the characteristics of Hearthstone that delineate the challenges that the game poses.


\subsection{Characteristics of Hearthstone}

As proposed by \citet{elias2012characteristics}, analyzing a game based on its characteristics is an important way to understand the challenges it poses. Some of the most salient characteristics of Hearthstone from the perspective of gameplaying agents are the following: the game has discrete inputs and outputs, meaning that the state observation is discrete and so is the action (i.e. what card to play and potentially its target). The observable game state has a natural and simple structured form (cards on the table and in hand), meaning that playing from pixels is unnecessary as it gives no new information and only adds an arbitrary computer vision problem. The branching factor is variable depending on game state, but generally high if one considers all the actions a player can take in a turn (five cards that can all be played in one turn, where each card can have one of five targets, lead to a branching factor of 375000; but it is not uncommon with game states where the player can choose only whether to play a single card or do nothing at all).

The game is considerably impacted by hidden information (partial observability), both in that the player does not know the opponent's hand and which cards will be drawn next from the deck. While some information can be observed or predicted (e.g. how long a particular card has been in an opponent's hand), only guesses can be made based on a priori knowledge of the deck or current known successful strategies. The game is also strongly impacted by stochasticity, both in the form of the initial shuffling of the decks and in the form of randomized effects of certain cards (for example, cards may deal variable damage or hit randomized enemies). The discrete and well-defined nature of the game makes it possible to build simulators of the game that can be executed at much faster than real-time, meaning that we can have fast forward models, which opens up for a large range of search- and planning-based methods for gameplaying. Finally, the deck is naturally separated into deck building, which nicely maps into what in most games is called strategy, and playing the decks, which we can call tactics. While these characteristics are the most salient for gameplaying agents, many of the challenges described in this paper are not primarily about playing the game (i.e. deciding which cards to play and when), so other characteristics will be discussed below.

The collectible card game Magic: The Gathering \citep{mtg:web19} shares many characteristics and challenges of Hearthstone: deck building, complex strategies, hidden information, large search space, etc. However this paper focuses on Hearthstone in particular because of its simplified mechanics and larger online player base. Such a player base currently means that there are more robust tools, simulators, and aggregated data like replays from  \textit{hsreplay.net}. While the higher complexity of mechanics in Magic: The Gathering poses interesting challenges, Hearthstone is a more accessible, state-of-the-art competitor. 



\subsection{Research Tools}

Hearthstone is supported by an active \emph{Hearthsim} community dedicated to building and maintaining simulators and other tools to help players strategize and study the game \footnote{https://hearthsim.info/}. Created and maintained primarily by darkfriend77, Sabberstone \footnote{https://github.com/HearthSim/SabberStone} is a fully functioning game simulator with at least fourteen contributors. The developers support research initiatives through their software and subreddit \footnote{A subreddit is a sub-forum on the website Reddit. Each subreddit is dedicated to a specific topic.}. Other simulators like Metastone \footnote{https://github.com/demilich1/metastone} and Spellsource \footnote{https://github.com/hiddenswitch/Spellsource-Server} have a fully functioning GUI for human players to play games.

\section{Playing the Game}\label{sec:playing}


Different games pose different challenges for playing. 
Some games are about long-term planning, while others are about quick reactions or estimating hidden information. The type of challenges posed to the gameplaying algorithm also depends on what kind of information and affordances are offered, e.g. whether information is presented as pure pixels or as information about objects, whether there is training time for the agent, and whether there is a fast forward model available. But the type of game-playing challenge offered also depends on how the game should be played. Apart from playing a game to win, there are other challenges, such as playing a game in a particular style, or creating heuristics that allow human players to learn to play the game by condensing knowledge about how to play it to a small set of rules.


\subsection{Playing to Win}

From a gameplaying perspective, Hearthstone offers a rare combination of challenges. It is a two-player, turn-based, and adversarial game, much like Chess, Go, and similar classic board games. However, like Poker it contains a substantial amount of hidden information; knowing which cards the opponent has in hand offers a considerable advantage, and good players spend significant effort trying to predict hands \citep{bursztein2016legend}. Tools like Predictor, which is a plugin for a third party data aggregator Hearthstone Deck Tracker \footnote{https://github.com/fatheroctopus/hdt-deck-predictor} exist to help players computationally determine these probabilities. 
Like many games, Hearthstone features stochasticity. An important source of stochasticity is the initial shuffling of the deck; some types of deck revolve around a particular card (e.g. Cthun, a very powerful high-mana card) which may by drawn early or late depending on the shuffling; additionally, many individual cards have stochastic effects (e.g. the Knife Juggler which randomly attacks one of the opponent's cards).

Most of the published academic work on Hearthstone to date focuses on methods for playing the game \citep{stiegler2017symbolic,swiechowski2018improving,santos2017monte,janusz2017helping,zhang2017improving}; in addition, there are a few papers about the closely related challenge of playing Magic \citep{ward2009monte}. Also, the several open-source simulators of Hearthstone mentioned previously are packaged with their own gameplaying agents. Most of the published work builds on Monte Carlo Tree Search (MCTS), a stochastic forward planning algorithm initially developed for Go but which has since seen much wider usage, and seeks to find ways the algorithm can be made to work with the game~\citep{santos2017monte,swiechowski2018improving,zopf2015comparison}. A key problem for tree search approaches is how to deal with that the agent does not know the opponent's hand. This missing information makes it impossible to expand the search tree based on the opponent's move to do a minimax search, unless a good guess of what their hand might be is available. Some of the work has therefore focused on learning predictive models of the opponent's hand~\citep{janusz2017helping,dockhorn2018predicting}. Other agents, such as that which is part of MetaStone, simply searches up until the end of the current move and uses a heuristic evaluation function, not even attempting to predict the opponent's move.

It is worth noting that all the published work on Hearthstone assumes that a fast simulator of some kind is available, which is an easy assumption to make because there are several. However, it is also possible to remove this assumption, and try to learn agents that play the game well without relying on search. This turns the problem into a reinforcement learning problem, where RL methods can potentially train neural networks that choose actions based on a representation of the current state. Such methods could potentially help developers with even quicker testing methods.

\subsection{Playing in Different Styles}

While the successful tactics of a Hearthstone player are at least partly determined by the deck, for many decks there are several different playstyles possible, and individual players will often prefer one playstyle over another. Can we create AI agents that can learn and recreate these playing styles, not only playing to win but doing so in the style of a particular player? This is a challenge that would seem to go above and beyond that of creating agents that ``simply'' play the game to win.

From a game design and development perspective there are several use cases for having agents capable of playing in specific styles. This includes providing examples to players of how to play as part of tutorials, offering interesting adversaries, and testing how some game design change will affect different types of players.

In the case of Hearthstone, perhaps the most important dimension of playing style variation is aggro-control. Playing ``aggro" means attacking the adversary early with all available resources, trying to decide the outcome of the game early. Playing ``control'' is a strategy for a longer game, where the player tries to stop the adversary from dominating the game while building up mana and card combinations for a late-game win. However, combo is a successful strategy that can take the form of aggro or control, but focuses primarily on combining the special effects of cards \citep{goes:tcaig17}.

One way of implementing agents with specific play styles is suggested by the \emph{procedural personas} concept. This entails modeling differences in play style as differences in objectives and search depth. In a previous application of this idea, personas in a roguelike dungeon crawler game were expressed as combinations of preferences for getting to the exit of a level fast, killing monsters, gathering treasure, and drinking health potions. By varying these preferences and evolving selection functions based on them, the same MCTS-based algorithm can be made to play in very different ways \citep{holmgaard2014evolving,holmgard2018automated}. The same approach could easily be implemented into Hearthstone by including varying preferences for winning early as opposed to late, using spells rather than minions etc.

\subsection{Finding Beginner Heuristics}

Another challenge which is related to, but not the same as, building or learning agents that can play Hearthstone is automatically finding human-teachable heuristics for playing it. In other words, using algorithms to find simple rules and strategies that can be communicated to humans in order to teach them how to play the game. Think of this as ``if you only had one rule for playing Hearthstone, what would it be?''. Maybe it would be something like ``if you have a minion and can attack the opponent's face directly do that; otherwise, attack another minion''. If you had two rules, what would they be?

As an example of an approach to this task, \citet{de2016generating} and \citet{de2018simplegenerating} developed a method for finding heuristics applied to the card games Blackjack and Texas Hold'em poker. In the most successful approach, genetic programming was used for finding lists of if-then rules (so called ``fast and frugal heuristics''). For BlackJack, it was found that as few as five rules could lead to almost-optimal play; these rules are much simpler to learn than the full strategy table~\citep{de2016generating}. In Texas Hold'em, a set of simple rules were discovered that led to at least novice-level play~\citep{de2018simplegenerating, de2018generating}.

What would beginner heuristics look like for Hearthstone? This is currently not known. We would need a description language that could capture relevant aspects of the game state as preconditions and relevant categories of in-game actions as consequences. This is an interesting research challenge that would teach us much about game design.

\subsection{Identifying Emergent Patterns}


Over time, the playerbase of some games develop a \emph{metagame}, which is often a collection of ideas about how to play the game well. It often includes a taxonomy around moves and action patterns that occur frequently in play, and ascribes varying degrees of strategic importance to them. An example in Chess is the Queen's Gambit, a centuries old opening move that is popular with players. By developing a shared taxonomy to describe patterns, both players and designers can better analyze, discuss, and evolve a metagame. 

Identifying emergent patterns of play and naming the most common or powerful requires human players to possess significant domain knowledge and experience like those developed for Chess \citep{gobet1996roles}, but AI and computational agents hold a significant advantage over these human analysts. While some human designers isolated common card combinations and plays in the puzzles released with the Boomsday expansion to Hearthstone, it should be possible to computationally analyze, identify, and categorize emergent patterns of gameplay. 

Hearthstone has two properties that suggest successful taxonomies can be created. First, it has a large and devoted playerbase, suggesting that any system developed to categorize play can be easily fact-checked with historical data and community knowledge like that available from \url{www.hearthscry.com/CollectOBot}. The key drawback to this is that games with large playerbases are likely to have quite thoroughly cartographed the action space, leaving only niche and novelty plays to be identified. Essentially, it presents the possibility that for many very popular games, this problem has been satisfactorily solved. However, Hearthstone's other key feature is that Hearthstone is a highly stochastic and evolving game. Blizzard regularly adds and removes cards to and from standard play, resulting in a dramatically and unpredictably shifting metagame every few months. The period immediately after the release of a new card set introduces a unique opportunity for AI to expedite the process of cartographing the game's newly reshaped action space.


Such work has several immediate short term applications and benefits. Identification of patterns of play is a key step in creating agents that mimic humans or take on specific play styles. Making human-recognizable moves with preexisting strategic connotations better enables both AI designers and the AI agents themselves to understand human play. For designers, this form of work will help identify possibly degenerate and undesirable behavior, like automated bots. As Hearthstone grows in popularity, and winning competitions becomes more valuable \citep{hodge2018business}, combating unfair play is becoming a more serious issue. However, sometimes strange player behavior could be due to an underlying issue in a game where players are somehow able to stall indefinitely, and AI can help identify what moves are leading to these unwanted board states. For players, identifying and categorizing plays can help to create better tutorials and automated trainers. With moves such as the Queen's Gambit, there are analyses dissecting when they are and are not appropriate. Using data to identify and analyze a single player's patterns can help to identify why they may be playing particularly well or poorly.

To the best of our knowledge, there is a limited body of work dedicated specifically to identifying known or recurring moves or patterns in games. However, this does not mean that no work has gone into the analysis and classification of moves. The identification of patterns of play is a problem that is closely related to the identification of player behavior. Where player behavior identification attempts to aggregate and analyze longitudinally along a single player's actions and moves, move pattern labelling aims to latitudinally observe plays across  many players to identify profiles that are more endemic to the design of the game or the playerbase at large. There has been substantial interest in player modelling, classification, and clustering in games. For example, work has been done to leverage player telemetry data to attempt to categorize and describe clusters of player behavior ~\cite{drachen2012guns}. 

\subsection{Difficulty Scaling}

Building an agent that plays the game well is an interesting problem in and of itself, but a high-performing agent does not necessarily lead to the most enjoyable user experience. Games like Hearthstone attract players of a wide variety of skill levels, and a game's far reaching appeal often depends on its ability to accommodate this variety of players. Games commonly offer AI opponents in some discrete set of difficulty levels (easy-hard, 1-10, etc), but the proposed values are ordinal, and may or may not be anchored against human player ability. Consistently providing players significant but not insurmountable difficulty and maintaining that challenge as they improve at the game is a non-trivial, but very important  \cite{apontescaling} problem for developers. This is a distinct problem from having agents that simulate different playstyles. The question is not whether an agent can play like a human, but rather whether they can play as well as or just barely better than a specific individual on command.
%
%

In a regular game of Hearthstone against an AI, both players have the same starting life and access to the same set of cards. There is no systemic or mechanical advantage one player has over the other. In this scenario, the challenge of tuning for difficulty lies solely in the implementation of the agent. Important questions to consider when approaching tunable difficulty in this case is how one would define a player-specific difficulty level in addition to how one would implement it. In addition to symmetric play, Hearthstone offers an asymmetric adventure mode similar to those in single-player role-playing games. In many single player games, it is possible to present challenge through asymmetry and some unfair advantage. In Hearthstone, this is manifested as unequal starting life totals or very powerful cards only usable by the opponent. The challenge of implementing these is shifted away from the agent's implementation and often toward tuning the severity of the asymmetry. Similarly, a developer might want to have access to a ``cheating'' bot for testing purposes, which is also a different problem than tuning for difficulty.

Work into difficulty scaling has explored a number of avenues toward modifying existing AI algorithms to scale difficulty as well as proposing novel agents that are scalable by design. For example, a technique known as Dynamic Scripting presents an algorithm with several parameters along which one could tune difficulty through manipulations of weights ~\cite{spronck2004difficulty}. Other work proposes the adaptation of concepts from psychological literature to measure and control the complexity of content created for single player games ~\cite{van2008difficulty}. There is little demonstrated work, however, toward difficulty scaling in symmetrical competitive games with well-established metrics for player skill.

\section{Building Decks} \label{sec:decks}

While playing the game given an existing deck has so far received the most attention, there are interesting challenges inherent in the domain of building these decks \citep{fontaine:gecco2019,bhatt:fdg18,garcia:cig16,garcia2018automated}. Human players often build them through experimentation and the evolving meta strategies of expert players, but automatically creating such decks could potentially lead to a richer diversity of meta strategies.  


Deciding on which cards to include in a deck often depends on the gameplay mode that a player selects. In the single player modes Tutorial, Adventure, and Missions, the cards are often chosen by the game designer. However, to maintain balance in the multi-player Play mode, players must choose a card format that dictates the types of cards in the decks. Wild cards include any of the over 1900 cards while Standard cards include the first two sets of cards (i.e. Basic and Classic) and any set released in the last two years. Often Standard decks can be composed to between six and eight sets of cards. Such rotation helps keeping the metagame from stagnating.

Another multi-player mode is Arena, where players build their decks one card at a time from a selection of three candidates (i.e. build their draft). Candidates are shuffled after every selection, meaning that players can but are not guaranteed to see their discarded choices again. While previous approaches to computational deckbuilding rely on a priori knowledge of the card pool and good card combinations, Arena mode forces players to make choices in real time without complete knowledge of the available cards. While some approaches could potentially help players draft cards like \citet{bursztein2016legend} who predicts cards an opponent will play based on replay data, and \citet{stiegler2016hearthstone} who develop a symbolic structure of cards, as of the time of writing there are currently no approaches specifically addressing Arena drafting.





\subsection{Transitivity and Dominance of Strategies}
Designers spend effort creating and balancing cards; when they introduce these new cards to the community it is important to maintain a degree of consistency with the old cards and decks while simultaneously facilitating the discovery of new deck archetypes. Complete transitivity of a deck space (i.e. a globally optimal deck) would quickly destroy properties of the game, and Hearthstone designers at Blizzard actively adjust properties of older cards or introduce new cards to ensure a variety of winning decks. However, some degree of transitivity is necessary for developing the art of deck building (i.e. randomly built decks should not be on average as good as those crafted with good strategies).

In Hearthstone cards are divided into subsets, currently including Basic, Classic, Expansion, and Adventure. Cards are mostly added through new sets in expansion and adventure. \citet{bhatt:fdg18} perform one of the first studies of the transitivity of the deck space of Hearthstone cards by holding playing strategies constant and looking only at the 133 cards available in the Basic set available to all players at the start of the game.  Preliminary results suggest some degree of transitivity in this space, but from the scope of the experiments the question remains to what degree these decks are transitive and if this transitivity is more or less present in different card sets.


\subsection{Deck Analysis: Mapping the Deck Strategy Space}

\begin{figure}
    \centering
    \subfigure[Space of Decks in an ES]
    {
        \includegraphics[scale=.2]{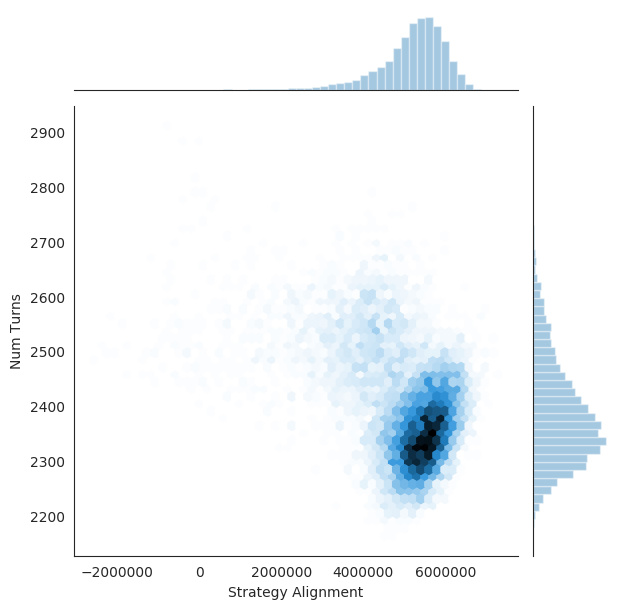}
    }
    \subfigure[Space of Decks in MAP-Elites]
    {
        \includegraphics[scale=.2]{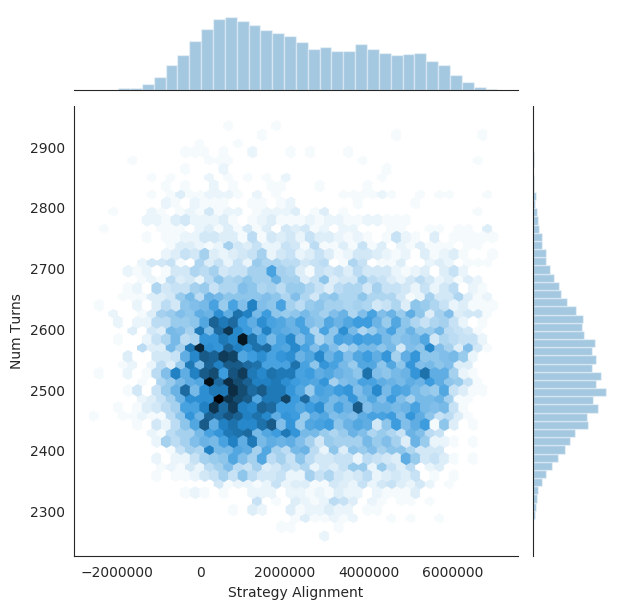}
    }
    \caption{\textbf{Example Search Spaces.} Distributions of ten thousand decks generated with a standard (1 + $\lambda$)-ES evolutionary strategy (shown in a) and MAP-Elites with Sliding Boundaries algorithm (shown in b) are plotted along two sample behavior metrics. Decks with the highest fitness are colored darker shades of blue. The x-axis is a measure called strategy alignment, which is a measure of how well the deck performance aligns with the player strategy. The y-axis is a measure of the number of turns taken in 200 hundred games. See \citet{fontaine:gecco2019} for more details.} \label{fig:searchspace}
\end{figure}

At the heart of evolutionary computation for deckbuilding is the idea that cards appearing in decks that perform well are more likely to be represented in future decks. Like many optimization algorithms including evolutionary strategies, search is biased toward one or several performance metrics. Examples of such performance metrics are winrate \citep{garcia:cig16} or the difference in health between players \citep{bhatt:fdg18,fontaine:gecco2019,silva:cog19}. 
In single objective optimization, the idea is that the search space will converge to a relatively small yet powerful space of near-optimal decks (shown in figure \ref{fig:searchspace}a).

While the convergence in figure \ref{fig:searchspace}a may at first seem to suggest the decks with higher x-values and lower y-values tend to perform best, figure \ref{fig:searchspace}b shows a variety of concentrations of highest-performing decks found with an algorithm that promotes deck diversity. Called MAP-Elites \citep{cully:nature15,mouret:arxiv15}, the algorithm builds a map of the best individuals and stores them if they also have unique behavior vectors. In figure \ref{fig:searchspace} the x-axis is a feature called strategy alignment and the y-axis represents the total number of turns over 200 games, but the number and type of these behaviors is theoretically limitless. Instead of investigating optimal behaviors of the high performing decks, we can look at how well decks tend to perform with different behaviors \citep{fontaine:gecco2019}.

\subsection{Deck Analysis: Identifying Cores, Weaknesses, and Strategy}

In collectible card games, it is common for players to build decks that contain particular core cards. Often, the rest of the cards in these decks support, strengthen, defend, or maximize the effectiveness of the core. In Hearthstone, examples of core cards include \emph{C'Thun}, which features a set of cards specifically designed to make \emph{C'Thun} stronger. A system capable of analyzing decks and identifying its cores is useful in validating a player's design around their preferred cards and generating more focused suggestions toward better executing on the deck's theme or core. Some tools like Archetypes \footnote{https://github.com/HearthSim/archetypes} developed by the Hearthsim community can help players identify such core cards and sets. However, in combination with powerful gameplaying strategies, such methods could be extended beyond what is currently popular in the metagame.

Another related topic for automated deck analysis involves identifying potential weakness or counters to a deck. In a robust metagame, no single card or deck will be entirely impervious to some sort of check or counter. For example, \emph{C'Thun} is a powerful core card. However, it can be countered by effects that force a player to discard \emph{C'Thun} without playing it. Such effects result in a great deal of wasted effort for the player. A system that can identify potential weaknesses to a deck or its core can be useful for systems that generate opponent decks against players or as tutorial generation systems to better enable players to learn about the intricacies of the game's competitive meta.

While there are some approaches to building decks compatible with a specific agent or playstyle \citep{bhatt:fdg18,garcia:cig16,garcia2018automated,fontaine:gecco2019}, fewer approaches identify the most effective agent, playstyle, or strategy given a specific deck. This may be useful when players or agents must play with decks they have limited control over, such as Tavern Brawl. This is particularly useful for less experienced players, who may be inclined to build decks with the starter cards they have, but lack an understanding of how to best play their cards. A very basic version of this feature can be found in several games. Yu-Gi-Oh! 5D's 2011 offered a very basic implementation which indicated what the playstyle of the deck appeared to be and identified several possible counters that players may need to account for.

\section{Assisting Players}\label{sec:assisting}

Another class of problems relate to building systems to assist players. While these problems are related and may be partly addressed with similar methods to AI for game playing and deck building, they are not identical. 

\subsection{Deck Building Assistance}

Building complete decks is an interesting AI challenge and can facilitate game testing, but most players need help completing decks. Assume a player has some favorite cards (e.g. \emph{Cthun}, \emph{Knife Juggler}, \emph{Hex}), and wants AI-based help finding a good balance of additional cards to make a complete and effective deck. One simple approach could be looking at the distribution of the mana costs of cards already included in the deck, and suggest new cards with the appropriate mana costs to balance the distribution; something like this recommendation system already exists in the Hearthstone client.
Other ad-hoc solutions\footnote{https://github.com/rembound/Arena-Helper} help players check card tier rankings for Arena Drafts\footnote{http://thelightforge.com/TierList}.
But those cards are not necessarily good complements to the cards already chosen. How could we do better?

Luckily, there are many approaches to recommendation items, perhaps because of its importance for e-commerce. Though the problems of recommending new books or clothes to buy for an online shopper differs from that of recommending cards for deck building, some approaches may be transferable; it is also likely that some algorithmic inventions from work on deck building assistance would carry over to other kinds of recommendation. As a potential starting point for a deck building assistance system, one might consider using the Apriori algorithm~\citep{agrawal1994fast} on a data set of high-quality decks, as collected from human players or deck-building algorithms \cite{fontaine:gecco2019}. By mining the co-occurrences of cards in high-quality decks, the Apriori algorithm would find association rules of the form ``if you have card A and card B in your deck, you might want to look at deck Q (because 34\% of decks that have cards A and B also have Q''). That is, if you like cards A and B, you may want to consider powerful decks with both. Now, it might be more useful to recommend a general class of cards rather than a particular class of cards and give a better explanation, such as ``you need a few minions with Taunt because you have many vulnerable minions''. What algorithm can give us this kind of advice? This is a fertile research problem.

\subsection{Gameplay Assistance}

The relationship of gameplay assistance to game-playing algorithms is similar to that of deck building assistance's relationship to deck building algorithms. What we are looking for here are systems that can help players play the game, for example by giving them feedback on how they are doing at the moment, suggesting what move to make next, or proposing a general strategy in response to observed play (e.g. ``the opponent seems to be going for rush, focus on taking out their minions''). 

One simple form of gameplay assistance could be a system that displays the winning probability at any given state, as calculated e.g. by Monte Carlo simulations, or by a win chance estimator trained on either simulations or logs of human games. Such a win chance estimator could also double as a state value function for a search-based approach to playing Hearthstone. The same approach, training on human or machine gameplay logs, would also work for constructing an action recommender. Win-rate predictors for Hearthstone have been the focus of previous research \cite{jakubik:fedcsis18}, driven specially by the AAIA Data Mining Competition \cite{janusz2018toward}. In general, a good starting point for this type of gameplay assistance systems would be a game-playing algorithm; a significant research challenge though, is which information to present to the player and how.

Another kind of gameplay assistance was suggested in the paper \emph{I am a legend: hacking Hearthstone using statistical learning methods}~\citep{bursztein2016legend}. 
By training on game logs, the system was able achieve high accuracy on predicting the next card played by the opponent, as high as 95\% on early game rounds and around 50\% in average. The system was labeled ``game breaking'' by Blizzard, and the creator agreed to not make his system publicly available. This raises questions about exactly how much, and which type, of gameplay assistance we want to have available.


\section{Assisting Designers} \label{sec:assistingdesigners}

Hearthstone is designed and developed by some of the world's foremost experts on online games, and in constant production by a team that intimately knows the game; new card sets, adventures and balance updates are regularly published. While it may be preposterous to suggest to creating new tools for assisting the designers that already know Hearthstone so well, the kind of challenges involved in designing, developing and producing such a game are similar to design and development tasks in many others. Therefore Hearthstone could be a versatile testbed for research on AI-assisted game design tools.

While the related research questions are less popular than game playing, there is active research on how to best use AI methods to assist game designers and developers. 
The paradigm of mixed-initiative co-creativity \citep{yannakakis2014mixed} envisions human designers creating games or other interactive systems in dialogue with an AI system. In it, both the human and the software can contribute and give feedback on what is being created.
Examples of systems attempting to enable mixed-initiative creativity in are platform game editors that can regenerate parts of the level while ensuring playability \citep{smith2011tanagra}, physics-based puzzle game editors that solve the puzzle for you \citep{shaker2013ropossum}, strategy game map generators that give feedback on balance and suggestions for how to improve maps \citep{liapis2013sentient}, and recommender systems that recommend new game elements based on machine-learned models of other games \citep{machado2016shopping} or the same game \citep{guzdial2018co}. The following sections describes how these approaches relate to generating cards in Hearthstone, balancing gameplay, and generating tutorials.

\subsection{Generating Cards}

Hearthstone has hundreds of different cards (over 1900), whereas Magic: The Gathering has more than ten thousand unique cards. Creating a genuinely new card (not essentially identical to some card that is already there), that adds value to the game, and that does not destroy the game's balance is a challenge. Could we use AI methods to help us here, for example through generating suggestions for new cards?
\citet{summerville2016mystical} describes such a system for Magic: The Gathering. The authors trained a sequence-to-sequence network on a large dataset of Magic cards, and use the trained model to generate new cards, including both statistics and descriptive text. The system produces cards that generally have recognizable statistics and grammatical descriptive text, but are often unbalanced, inconsistent or in other ways game-breaking. But in many cases, these cards can easily be turned into playable cards with a little human intervention.

\citet{ling:arxiv16} focus on the problem of generating valid code from natural language descriptions on Hearthstone and Magic cards. This is in a sense the other side of the coin, and would be needed as a part of a functioning card generation system, to use the code for these decks for artificial agents that can playtest new cards.

\subsection{Balancing the Game}

Creating a card is one thing, but it is difficult to create a new card while maintaining balance in gameplay. Many cards that seem to have reasonable mana cost and attack and defense values may, when combined be certain other cards, enable game-breaking strong combos. Balancing a collectible card game is a major undertaking, and is typically done manually through extensive human playtesting and in response to observed player behavior. It could be seen as an optimization process, where the desired outcome is to have a reasonably low range of usefulness of individual cards, or alternatively to have a large range of useful card combinations and low variance between the value of these. However, formulated in this way, the optimization problem is almost certainly untractable for games with large deck spaces such as Hearthstone \cite{garcia2018automated}. For somewhat simpler card-based games such as Dominion, evolutionary approaches to this problem can work well, as demonstrated by \citet{mahlmann2012evolving}. In Hearthstone, a feasible approach to computational balancing might be to search for decks that involve the proposed new card and nerf (lower the stats of) the card if a deck is found that is too strong.

\subsection{Generating Tutorials}

Tutorials are an important part of modern video games, which are primarily learned through play. However, constructing effective and accessible tutorials is complex and labor-intensive, and AI methods could help reduce this burden for game developers. Some recent approach address automatically constructing game tutorials. For example, the \emph{AtDelfi} system analyzes mechanics of arcade-style games to generate videos and written instructions that instruct players to play the game~\cite{green2018atdelfi}. Hearthstone could serve as a suitable testbed for algorithms that generate challenge problems that teach you the mechanics of the game and useful heuristics, or demonstrations of such mechanics and heuristics. Examples of desirable results would be the introductory puzzles present in the card games Eternal~\cite{eternal:web19} and Faeria~\cite{faeria:web19}.


\section{Conclusion}

We have described a large number of research and application challenges for AI arising from a single game, Blizzard's Hearthstone. These applications span almost the entire field of artificial intelligence and games and the reader may wish to compare the particular challenges to those outlined by~\citet{yannakakis:book18}. At the same time, the nature of this particular game (such as its hidden information, adversarial nature, discrete state and action space, high and variable branching factor, stochasticity, relative ease of forward modeling, and separation of deck building from playing) shape the particular form of the challenges it poses. As such, the challenges are rather different from those posed by some of the dominating game-based AI benchmarks, such as Atari/ALE, Doom, Chess, and Go. Notably the dominant approaches for playing these games (e.g. training deep neural networks with reinforcement learning to play based on pixel inputs, or searching ahead in the game tree with MCTS) are missing from the suggestions above. While it might be possible to use Deep Q-learning to learn to play based on pixel inputs, it would be complicating matters so much as to almost be nonsensical when much better representations of the game state is available; and it is very hard to use MCTS effectively beyond a single turn given the partial observability of the game. But the diverse challenges posed by Hearthstone are no less interesting from an AI perspective. This underscores the need to choose the game to use for your AI benchmark carefully, and play and think about that game to understand the challenges it poses. 

\begin{acknowledgements}
Thanks to Matthew Fontaine, Aditya Bhatt, Connor Watson, Param Trivedi for their contributions to research that has informed this paper. Additionally, we thank Andy Nealen and Alex Zook for useful discussions. Finally, we are happy that all the hours we spent playing the game could contribute to something useful, or at least publishable.
\end{acknowledgements}

\bibliographystyle{apalike}
\bibliography{ki.bib}

\end{document}